\documentclass[11pt,a4paper]{article}
\usepackage[hyperref]{acl2020}
\usepackage{times}
\usepackage{latexsym}
\usepackage{amsfonts}
\usepackage{graphicx}
\usepackage{latexsym}
\usepackage{booktabs}
\usepackage{amssymb}
\usepackage{amsmath}
\usepackage{pifont}
\usepackage{url}

\usepackage{bm}
\usepackage{multirow}
\usepackage{pstool}

\DeclareMathOperator*{\Sigmoid}{Sigmoid}
\DeclareMathOperator*{\Gate}{Gate}
\DeclareMathOperator*{\MultiHead}{MultiHead}

\usepackage{microtype}

\aclfinalcopy

\title{Let's be Humorous: Knowledge Enhanced Humor Generation}
\author{
Hang Zhang, Dayiheng Liu, Jiancheng Lv\hspace{1mm}\thanks{\hspace{2mm}Correspondence to Jiancheng Lv.},Cheng Luo \\
\hspace{0.5mm} College of Computer Science, Sichuan University\\
{\tt  zhanghang.scu@gmail.com}
 \\
 {\tt losinuris@gmail.com}\\
 {\tt lvjiancheng@scu.edu.cn}\\
  {\tt wulaoshi\_luocheng@foxmail.com}
  } 

\date{}
\begin{document}
\maketitle
\begin{abstract}
The generation of humor is an under-explored and challenging problem. Previous works mainly utilize templates or replace phrases to generate humor. However, few works focus on freer forms and the background knowledge of humor. The linguistic theory of humor defines the structure of a humor sentence as set-up and punchline. In this paper, we explore how to generate a punchline given the set-up with the relevant knowledge. We propose a framework that can fuse the knowledge to end-to-end models. To our knowledge, this is the first attempt to generate punchlines with knowledge enhanced model. Furthermore, we create the first humor-knowledge dataset. The experimental results demonstrate that our method can make use of knowledge to generate fluent, funny punchlines, which outperforms several baselines. Our data and code are publicly available at \url{ https://github.com/onedoge/Knowledge-Enhanced-Humor-Generation}.

\end{abstract}

\section{Introduction}
Humor is prevalent in daily communication and often expresses highly developed human knowledge and emotion. However, the automated generation of humor has always been a great challenge, which requires not only a deep understanding of the semantic but also a full consideration of cultural background. 

\begin{figure}
   \centering
   \includegraphics[width=0.45\textwidth]{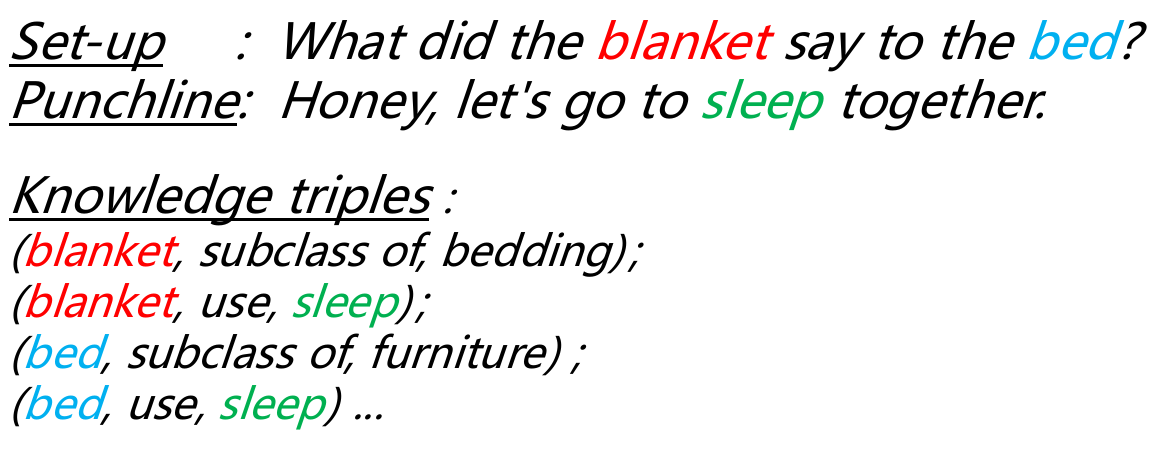}
   \caption{An illustration of a satirical joke. The set-up sentence provides the context and the punchline produces a laugh. The knowledge is organized in triple-types which is helpful to understand this joke. 
   }
   \label{fig:fig_case}
\end{figure}

Jokes are the primary carrier of humor. According to the Inconsistency Theory, a joke generally consists of set-up and punchline~\citep{bright1992international}. Consider the example in Fig.~\ref{fig:fig_case}: `` What did the blanket say to the bed?". The question, which is also the set-up, provides the context for this joke. The punchline, ``Honey, let's go to sleep together.", is usually at the end of a joke and produces a laugh.

Previous methods for humor generation have been mainly based on fixed templates or lexical substitution~\citep{petrovic2013unsupervised,DBLP:conf/acl/ValituttiTDT13,DBLP:conf/emnlp/HossainKVHK17,DBLP:conf/acl/WanTY18}. Due to lack of context, they can only produce generic and isolated jokes. Besides, background knowledge is crucial in understanding and generating jokes. In the above example, if we don't know the background of both entities, we wouldn't feel the humor from this joke. However, as far as we know, the background knowledge of jokes has not been introduced in the current computational humor research.

As mentioned above,  we propose the task of generating punchlines with the set-up and relevant knowledge. For this task, we create the first dataset that contains set-ups, punchlines and background knowledge. Furthermore, we propose a framework as shown in Fig.~\ref{fig:fig1}. The relative background is converted into a knowledge graph and encoded by our proposed knowledge encoder. When generating the punchline, the decoder will first attend to the information from the set-up encoder, then fuse knowledge representation by knowledge fusion layer. The experiments indicate that our model performs better than strong baselines and can generate funny punchlines.

Our contributions are threefold: (1) We make the first attempt to generate punchline with the set-up and relevant knowledge. (2) We propose a framework to integrate external knowledge into end-to-end generation framework. (3) We provide the first dataset of knowledge paired with jokes for further study.

\section{Related Work}

\noindent\textbf{Humor Theory} Incongruity theory has an essential guiding position in the field of computational humor~\citep{binsted2006computational}. It believes that the inconsistency between the reader's expectation and the ending of one story is the key to humor generation and verbal irony~\citep{AmirWLCS16}. On this basis, SSTH (Script-based Semantic Theory of Humor) theory is proposed~\citep{raskin2012semantic}. SSTH defines the structure of a joke as set-up and punchline. The set-up provides humorous context information, including multiple possible explanations (scripts). The punchline, usually at the end of a joke, points to a surprising explanation that triggers the humorous effect. According to this theory, we explore how to generate a punchline given the set-up sentence.

\noindent\textbf{Humor Generation} ~\citet{petrovic2013unsupervised} attempt to fill in the blank of the fixed template ``I like my \textbf{X}, like I like my \textbf{Y}, \textbf{Z} " in an unsupervised way with four customized hypotheses. \citet{DBLP:conf/acl/ValituttiTDT13} substitute words with taboo words to generate adult jokes. \citet{DBLP:conf/emnlp/HossainKVHK17} use a classifier to help people choose humorous words in a fill-in-the-blank~\cite{liuetal2019tigs} game. \citet{DBLP:conf/acl/WanTY18} encode multiple meanings of a word and use a hybrid beam search method to generate puns. \citet{heetal2019pun} propose a retrieve-and-edit pipeline to generate a pun sentence. Different from these work, we consider the punchline generation with the use of world knowledge.

\section{Humor Corpus with Background Knowledge}

To prepare our dataset, we choose Short Jokes dataset\footnote{\url{https://www.kaggle.com/abhinavmoudgil95/short-jokes}} and Reddit-Joke dataset\footnote{\url{https://www.kaggle.com/cuddlefish/reddit-rjokes}} as raw data, which are public on Kaggle. Then we perform joke filtering, punchline segmentation, and joke de-duplication. We first remove the data that contains the special characters and only keep the jokes with at least two sentences and fifteen words. Then we treat the last clause of the joke as punchline and the rest as set-up. For de-duplication, we use the BOW (bag of words) and cosine similarity to detect the sentence similarity. Jokes with similarity greater than 0.93 are deleted.
\begin{figure}
   \centering
   \includegraphics[width=0.5\textwidth]{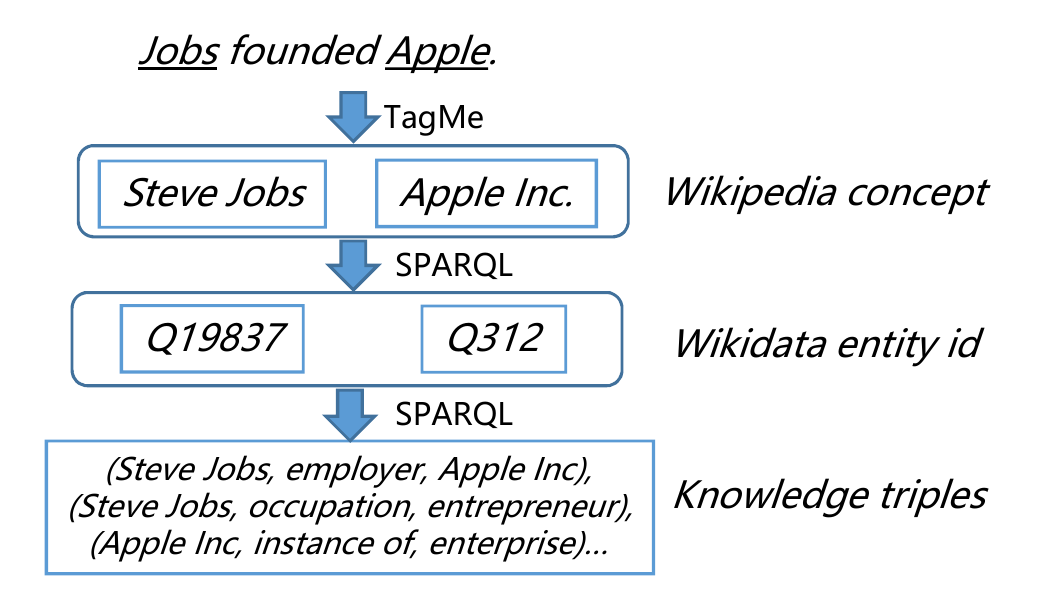}
   \caption{An example of knowledge acquisition for the sentence ``Jobs founded Apple.". Firstly, we use TagMe for entity linking. After getting the Wikipedia concepts, We obtain knowledge triples through Wikidata Query Service.
   }
   \label{fig:fig_data}
\end{figure}

To obtain background knowledge of the set-up sentences, we use the entity link tool TagMe~\citep{DBLP:conf/cikm/FerraginaS10}. As an example shown in Fig.~\ref{fig:fig_graph}, TagMe can map entities in sentences to concepts in Wikipedia and give confidence of the mapping. To ensure the credibility of entities, we only keep entities with confidence greater than 0.1. After getting the entity's concepts on Wikipedia, we use SPARQL to link entities to Wikidata and get the entity-related triples. Overall, our dataset contains about 107,000 data pairs. We divide the training set, verification set and test set according to the 7:2:1 ratio.

\begin{figure*}[ht]
   \centering
   \includegraphics[width=0.98\textwidth]{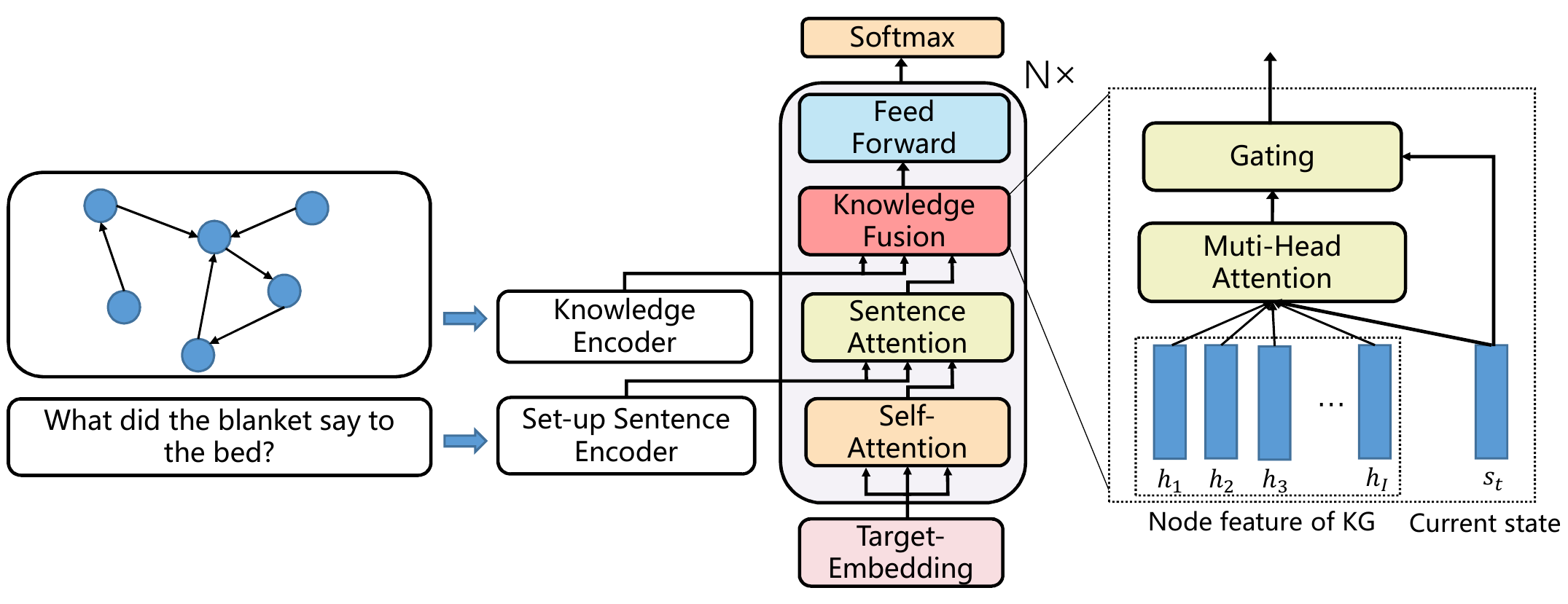}
   \caption{The overview of the proposed framework, which consists of a knowledge encoder, a set-up sentence encoder and a decoder with knowledge fusion layer. 
   }
   \label{fig:fig1}
\end{figure*}
\section{Methodology}

\subsection{Problem Definition and Model Overview}
We formulate the task of punchline generation with the set-up and relative knowledge. One knowledge triple is composed of subject $s$, relation $r$ and object $o$, denoted as $ k = (s,r,o)$.  Given a set-up sentence $\mathbf{X}=\left\{x_{1}, x_{2}, \ldots, x_{p}\right\}$ and its background knowledge triples $\mathbf{K}=\left\{k_1,k_2, \ldots,k_u\right\} $, our goal is to generate a punchline $\mathbf{Y}=\left\{y_{1}, y_{2}, \ldots, y_{q}\right\}$. 

Our model is based on Transformer~\citep{vaswani2017attention}. The overview is shown in Fig.~\ref{fig:fig1}. Compared with the origin Transformer structure, we add two modules, knowledge encoder and knowledge fusion layer. Knowledge encoder obtains the hidden features of background knowledge. The knowledge fusion layer fuses knowledge features into the decoding process after the multi-head attention layer in the decoder. 
\begin{figure}
   \centering
   \includegraphics[width=0.4\textwidth]{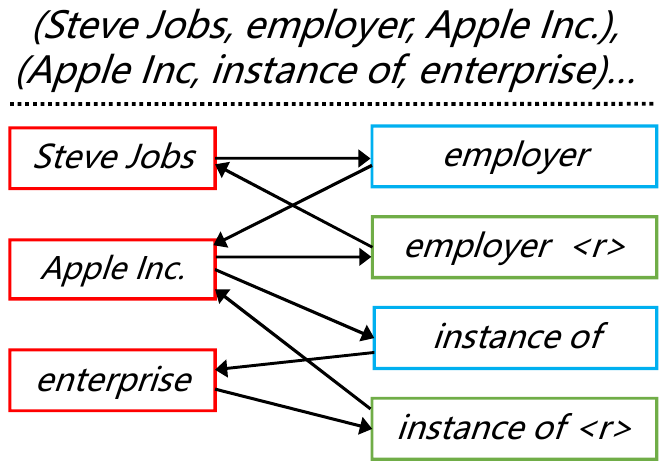}
   \caption{An example of constructing knowledge graph. Top: knowledge triples. Bottom: knowledge graph. The red, blue, and green boxes represent entity, forward relation, and reverse relation nodes, respectively. The reverse relation nodes are identified by the symbol $<r>$.
   }
   \label{fig:fig_graph}
\end{figure}

\subsection{Constructing Knowledge Graph}
Given a knowledge triple set $\mathbf{K}=\left\{k_1,k_2, \ldots,k_u\right\}$, we turn it into a directed graph. An example is shown in Fig.~\ref{fig:fig_graph}. Specifically, the co-referential entities in set $\mathbf{K}$ are folded into a single entity node, and the relations are mapped into relation nodes (The entity here is the subject and object). The subject, relation, and object nodes in one triple are connected in turn. In order to allow the information of the object to flow into the subject node, we add a reverse relation node which is similar to~\citet{DBLP:conf/naacl/Koncel-Kedziorski19}. Since entities and relationships in Wikidata are usually multi-word expressions, we encode these words with Bi-directional Long-Short Term Memory (Bi-LSTM)~\cite{hochreiter1997long,schuster1997bidirectional}. We adopt the last hidden states as the initial features of nodes. Finally, we get a connected graph $\mathbf{G} = (\mathbf{V}, \mathbf{E} ,\mathbf{H}^0)$, where $\mathbf{V}$ is the set of nodes, $\mathbf{E}$ is the set of edges, $\mathbf{H}^0$ is the initial feature set of $\mathbf{V}$.
\subsection{Knowledge Encoder}
We use the  graph attention network~\citep{DBLP:conf/iclr/VelickovicCCRLB18} to incorporate the features of adjacent nodes in $\mathbf{G}$. For a knowledge graph $\mathbf{G} = (\mathbf{V}, \mathbf{E} ,\mathbf{H}^l), \mathbf{V}=\left\{{v}_{1}, {v}_{2}, \ldots, {v}_{I}\right\}, \mathbf{H}^l=\left\{{h}_{1}^l, {h}_{2}^l, \ldots, {h}_{I}^l\right\}$, the initialization feature of node  $v_i$ is $h_i^{0}$. Each node updates its feature through $M$-headed self-attention by receiving information from its neighbors, which can be described as follows.
\begin{align}
&h_{i}^{(l+1)}=\|_{m=1}^{M} \sigma\left(\sum_{j \in \mathcal{N}(i)} \alpha_{i j}^{m} \mathbf{W}_{V}^{m} h_{j}^{l}\right),\\
&\alpha_{i j}^{m}=\frac{\exp \left(\left(\mathbf{W}_{K}^{m} h_{j}^{l}\right)^{\top} \mathbf{W}_{Q}^{m} h_{i}^{l}\right)}{\sum_{j \in \mathcal{N}_{(i)}} \exp \left(\left(\mathbf{W}_{K}^{m} {h}_{j}^{l}\right)^{\top} \mathbf{W}_{Q}^{m} {h}_{i}^{l}\right)},
\label{eq:2}
\end{align}

where the feature of node $i$ in layer $l$ is $h_{i}^{l}$, $h_{i}^{l} \in \mathbb{R} ^{d}$. $M$ is the number of heads, $\|$ denotes the concatenation of $M$ attention heads. $\mathcal{N}(i)$ is all one-hop neighbors of $v_i$ (include $v_i$), and $\sigma$ is an activation function. $\mathbf{W}_{Q}^{m}, \mathbf{W}_{K}^{m},\mathbf{W}_{V}^{m}\in \mathbb{R} ^{d \times (d/M)}$ map $h_{i}^{l}$ and $h_{j}^{l}$ to the $m$-th head subspace, and we calculate the connection score $\alpha_{i j}^{m}$ by Eq. (\ref{eq:2}). 

\subsection{Decoder with Knowledge Fusion Layer}
Before the knowledge fusion layer, the decoder's operation is the same as the original Transformer. Assume that the feature of nodes obtained by the knowledge encoder is $\mathbf{H} = \left\{h_1, h_2, \ldots ,h_I\right\}$, and the input sequence of the decoder at time $t$ is $\mathbf{Y}_t = \left\{y_0, y_1,\ldots,y_t\right\}$. We use a stack of $N$ identical blocks to compute target-side representations. Each block is composed of four sub-layers as shown in Fig.~\ref{fig:fig1}. In $n$-th block, after the masked multi-head attention calculation with set-up sentence, the hidden state is expressed as $\mathbf{S}^n = \left\{s_1^n, s_2^n, ... s_t^n\right\}$. 

The knowledge fusion layer contains a multi-head attention layer~\citep{vaswani2017attention} and a gating machine inspired by highway network~\citep{srivastava2015training}. Firstly, we integrate knowledge feature into the current state. 
\begin{equation}
\mathbf{A}^n = \MultiHead(\mathbf{S}^n,\mathbf{H},\mathbf{H}).
\end{equation}
Note that the node information in the background knowledge graph may contain noise due to the inaccuracy of entity link tools. To address this problem, we introduce the gating mechanism to allow for a better trade-off between the impact of background knowledge and the information from set-up encoder.
\begin{equation}
\Gate(\mathbf{S}^n) = \lambda^n \mathbf{S}^n +(1-\lambda^n) \mathbf{A}^n,
\end{equation}
where $\lambda$ denotes the gating weight, which is given by
\begin{equation}
\lambda^n = \Sigmoid(\mathbf{W}_g^n\mathbf{S}^n),
\end{equation}
where $\mathbf{W}_g$ is a model parameter.

Then we input the feature to the feed-forward layer of the Transformer. After the operations of $N$ blocks, we get the final state $\{e_1,e_2,\ldots,e_t\}$. Finally, the probability distribution of generating the next target word $y_{t+1}$ can be expressed as:
\begin{equation}
P\left(y_{t+1} | \mathbf{X}, \mathbf{K}, y_{<=t} ; \theta\right) \propto \exp \left(\mathbf{W}_{o} e_{t}\right),
\end{equation}
where $\mathbf{W}_0 \in \mathbb{R}^{\left|\mathcal{V}_{y}\right| \times d}$ is a model parameter, $\left|\mathcal{V}_{y}\right|$ is the target vocabulary size.

\section{Experiments}

\subsection{Evaluation}
We use ROUGE-1, ROUGE-2, and ROUGE-L as automatic evaluation metrics, which measure the similarity between the output and the reference. We also conduct a human evaluation. For each model, we randomly select 40 set-ups and relative knowledge from the test set to generate punchlines. For further comparison, we also involve 40 human-written jokes. We invite 5 evaluators who are good at English and have the proper world knowledge to rate these jokes.  We set three standards for evaluators to judge the punchlines: (1) Grammar and fluency (Is the punchline written in well-formed English?); (2) Coherency (Is the punchline coherent with the set-up sentence?); (3) Funniness (Is the punchline funny?). The score of each aspect ranges from 1 to 5, with the higher score the better. 
\begin{table*}[tb]
	\centering
	
    \resizebox{\textwidth}{!}{
    \begin{tabular}{ll}
    \hline
        \textbf{Set-up} & Trump wants to cut funding for birth control, renegotiate trade deals and stop the wars. \\ \hline
        \textbf{Knowledge}   & \begin{tabular}[c]{@{}l@{}}(Donald Trump, field of work, politics); (Donald Trump, position held, President of \\the United States);  (Birth control, part of, human population planning) \ldots\end{tabular}  \\\hline
        \textbf{S2S-GRU} & He is not in denial.\\ 
        \textbf{CopyNet} & It was not a solution.\\ 
        \textbf{Transformer} & They are making headlines. \\
        \textbf{Our model} &  It seems he is a really nice president. \\ 
        \textbf{Human-written} &  It seems pulling out is his solution for everything. \\ \hline              
        \hline
        \textbf{Set-up} & Cocaine makes people happy, what's the most dangerous thing about it? \\ \hline
        \textbf{Knowledge} &  \begin{tabular}[c]{@{}l@{}} (Cocaine, instance of, 
        drug); (Cocaine,medical  condition treated, pain); \\ (Cocaine, Description, strong stimulant used as a recreational drug) \ldots\end{tabular}  \\\hline
        \textbf{S2S-GRU} & Be important.\\ 
        \textbf{CopyNet} & Happy life with danger.\\ 
        \textbf{Transformer} & It seems to be more safe. \\
        \textbf{Our model} &  It is a drug. \\ 
        \textbf{Human-written} &  Maybe getting caught by the police \\\hline 
    \end{tabular}}
    \caption{Example outputs with four different models. Our model can generate more coherent  punchlines.  }
     \label{table:case}
\end{table*}
\subsection{Baselines and Implementation Details}
Since there is no direct related work of this task, we compared three widely used text generation methods, including S2S-GRU with attention mechanism~\citep{DBLP:journals/corr/BahdanauCB14}, CopyNet~\citep{DBLP:conf/acl/GuLLL16}, and Transformer~\citep{vaswani2017attention}. By comparing these models, it is shown that our method can use knowledge to enhance punchline generation.

In pre-processing, we use the pre-trained BPE dictionary with the vocabulary size of 25000 from~\citet{heinzerling2018bpemb}. Knowledge encoder uses 2 layers. For Transformer baseline and our model, the embedding and the hidden dimensions are 512. The block number of encoder and decoder is set to 4, the number of attention heads is set to 8, and the size of feed-forward layers is set to 2048. For S2S-GRU and CopyNet, the embeddings of words are 256 dimensions. We use 1-layer bidirectional GRU~\citep{choetal2014learning} with the hidden size of 256 as encoder. The decoder is a 2-layer GRU with the hidden size of 256. We also tried to increase the number of parameters of the S2S-GRU and CopyNet, but we did not get better results.

For our model, two-step training strategy is employed, inspired by \citealt{DBLP:conf/emnlp/ZhangLSZXZL18}. Specifically, we first pre-train a standard Transformer, which is used to initialize the parameters of the  set-up encoder and partial decoder. Then we fine-tune  the entire model. During training, we use the Adam optimization~\citep{DBLP:journals/corr/KingmaB14} with 16 mini-batch size. The learning rate is set to 0.001. During decoding, we implement beam search with beam size 5. 
\begin{table}
\begin{center}
\setlength{\tabcolsep}{1mm}{
    \begin{tabular}{c|c|c|c}
     \hline 
        Method & ROUGE-1 & ROUGE-2 & ROUGE-L\\
     \hline
        S2S-GRU & 22.79 & 5.35 & 19.85 \\
        CopyNet & 22.31 & 4.66 & 20.29 \\
        Transformer & 23.73 & 6.27 & 21.89 \\
     \hline
        Our model  & 25.97 & 9.47 & 23.60   \\
    \hline
    \end{tabular}}
 \end{center}
    \caption{Automatic evaluations of generation models.}
    \label{table:rouge}
\end{table}
\begin{table}
\setlength{\tabcolsep}{1mm}{
    \begin{tabular}{c|c|c|c}
     \hline 
        Method & Fluency & Coherency & Funniness\\
     \hline
        S2S-GRU & 2.84 & 2.02 & 2.16 \\
        CopyNet & 2.60 & 2.48 & 2.04 \\
        Transformer & 3.04 & 2.86 & 2.40 \\
        Our model & 3.24 & 3.28 & 2.60 \\
     \hline
        Human-written & 4.06 & 3.88 & 3.30   \\
    \hline
    \end{tabular}}
    \caption{ Human evaluation of generation models.}
    \label{table:human}
\end{table}
\subsection{Result}
Tab.~\ref{table:rouge} shows the automatic evaluation results. We find that: (1) As expected, Transformer based methods perform better than other baselines. 
(2) Background knowledge can promote punchline generation, by comparing our method with origin Transformer.

Human evaluation results are shown in Tab.~\ref{table:human}. Our method performs better than three baselines in all metrics. These results demonstrate the effectiveness of our model with knowledge enhanced. Nevertheless, there is still a  gap between generated punchlines and expert-written punchlines across all aspects, indicating that humor generation remains an open challenge. Interestingly, the funniness score of human-written jokes is not very high, due to different people's sensitivity to humor. It is consistent with~\citet{petrovic2013unsupervised}.

\subsection{Case Study}
Tab.~\ref{table:case} shows examples of various model outputs for two particular test instances. In general, all models produce fluent punchlines, but the semantic coherency between generated punchlines and set-up sentences is poor. Compared with baselines, our proposed method can generate more fluent and coherent punchlines. The first example is related to political commentary, which is often satirical humor. To some extent, the outputs of our method contains the background information of ``Trump" and ``Cocaine" .

\section{Conclusion and Future Work}

In this paper, we make the first endeavor to generate punchline, a freer form of humor generation. Besides we propose a knowledge-enhance framework which is generic and novel. Experiments show our method can make use of knowledge to enhance punchline generation. Future work can improve the knowledge selection method and add the explicit features of humor to the model.

\section{Acknowledgement}
This work is supported by the National Key R\&D Program of China under contract No. 2017YFB1002201, the National Natural Science Fund for Distinguished Young Scholar (Grant No. 61625204), and partially supported by the Key Program of National Science Foundation of China (Grant No. 61836006).

\bibliography{anthology,acl2020}
\bibliographystyle{acl_natbib}

\end{document}